\pdfoutput=1
\documentclass[11pt]{article}

\usepackage[]{ACL2023}

\usepackage{times}
\usepackage{latexsym}

\usepackage[T1]{fontenc}
\usepackage[utf8]{inputenc}

\usepackage{microtype}

\usepackage{inconsolata}
\usepackage{array}

\usepackage{natbib}

\usepackage{graphicx}
\usepackage{subcaption}
\usepackage[normalem]{ulem}

\usepackage{tikz}
\usetikzlibrary{arrows}
\usepackage{tikz}

\usepackage{tikzscale}
\usetikzlibrary{arrows.meta,
                calc,
                positioning,
                quotes,
                shapes.geometric}

\pgfdeclareshape{datastore}{
  \inheritsavedanchors[from=rectangle]
  \inheritanchorborder[from=rectangle]
  \inheritanchor[from=rectangle]{center}
  \inheritanchor[from=rectangle]{base}
  \inheritanchor[from=rectangle]{north}
  \inheritanchor[from=rectangle]{north east}
  \inheritanchor[from=rectangle]{east}
  \inheritanchor[from=rectangle]{south east}
  \inheritanchor[from=rectangle]{south}
  \inheritanchor[from=rectangle]{south west}
  \inheritanchor[from=rectangle]{west}
  \inheritanchor[from=rectangle]{north west}
  \backgroundpath{
    \southwest \pgf@xa=\pgf@x \pgf@ya=\pgf@y
    \northeast \pgf@xb=\pgf@x \pgf@yb=\pgf@y
    \pgfpathmoveto{\pgfpoint{\pgf@xa}{\pgf@ya}}
    \pgfpathlineto{\pgfpoint{\pgf@xb}{\pgf@ya}}
    \pgfpathmoveto{\pgfpoint{\pgf@xa}{\pgf@yb}}
    \pgfpathlineto{\pgfpoint{\pgf@xb}{\pgf@yb}}
 }
}

\usepackage{adjustbox}
\usepackage{array} % Required for new column type

\newcolumntype{C}[1]{>{\centering\arraybackslash}m{#1}}

\makeatletter
\def\@fnsymbol#1{}
\makeatother

\title{
Psychological Profiling in Cybersecurity: A Look at LLMs and Psycholinguistic Features
}

\author{\fontsize{10}{10}\selectfont Jean Marie Tshimula$^{1,2,3}$ D'Jeff K. Nkashama$^{1,3}$ Jean Tshibangu Muabila$^{1,4}$ René Manassé Galekwa$^{1,2,5}$ \\ {\fontsize{10}{10}\selectfont \bf  Hugues Kanda$^{1}$ Maximilien V. Dialufuma$^{1,6}$  Mbuyi Mukendi Didier$^{1,2,7,8}$ Kalonji Kalala$^{1,9}$} \\ {\fontsize{10}{10}\selectfont \bf  Serge Mundele$^{1}$ Patience Kinshie Lenye$^{1}$ Tighana Wenge Basele$^{1,2,10}$ Aristarque Ilunga$^{1,2}$} \\ {\fontsize{10}{10}\selectfont \bf Christian N. Mayemba$^{1}$ Nathanaël M. Kasoro$^{2}$ Selain K. Kasereka$^{2}$ Hardy Mikese$^{11}$}\\ {\fontsize{10}{10}\selectfont \bf Pierre-Martin Tardif$^{3}$ Marc Frappier$^{3}$ Froduald Kabanza$^{3}$ Belkacem Chikhaoui$^{12}$ } \\ \bf {\fontsize{10}{10}\selectfont Shengrui Wang$^{3}$ Ali Mulenda Sumbu$^{13}$ Xavier Ndona$^{14}$ Raoul Kienge-Kienge Intudi$^{15}$}  
\thanks{$^1$Groupe de Recherche de Prospection et Valorisation des Données (Greprovad), Global $^{2}$Department of Computer Science, University of Kinshasa, DRC $^{3}$Department of Computer Science, Université de Sherbrooke, Canada $^{4}$LISV-UVSQ, Université Paris-Saclay, France $^{5}$University of Klagenfurt, Austria $^{6}$Montreal Behavioural Medicine Centre, Centre Intégré Universitaire de Santé et Services Sociaux du Nord-de-l’Île-de-Montréal (CIUSSS-NIM), Canada $^{7}$Biomedical Research Unit, Hospital Monkole, Kinshasa, DRC $^{8}$University of Florida, USA $^{9}$School of Electrical Engineering and Computer Science, University of Ottawa, Canada $^{10}$Karlstad University, Sweden $^{11}$Institut Supérieur Pédagogique de Kikwit, DRC $^{12}$Applied Artificial Intelligence Institute, TELUQ University, Canada $^{13}$Faculty of Psychology and Education Sciences, University of Kinshasa, DRC $^{14}$Harrisburg University of Science and Technology, USA $^{15}$School of Criminology, University of Kinshasa, DRC. Correspondence email: \href{mailto:jeanmarie.tshimula@unikin.ac.cd}{\texttt{jeanmarie.tshimula@unikin.ac.cd}} and \href{mailto:nkad2101@usherbrooke.ca}{\texttt{nkad2101@usherbrooke.ca}}}
} 

\date{}

\begin{document}

\maketitle

\begin{abstract}
The increasing sophistication of cyber threats necessitates innovative approaches to cybersecurity. In this paper, we explore the potential of psychological profiling techniques, particularly focusing on the utilization of Large Language Models (LLMs) and psycholinguistic features. We investigate the intersection of psychology and cybersecurity, discussing how LLMs can be employed to analyze textual data for identifying psychological traits of threat actors. We explore the incorporation of psycholinguistic features, such as linguistic patterns and emotional cues, into cybersecurity frameworks. Our research underscores the importance of integrating psychological perspectives into cybersecurity practices to bolster defense mechanisms against evolving threats.

\end{abstract}

\section{Introduction}

{\color{black}

Psychological profiling plays a crucial role in cybersecurity, particularly in understanding and identifying the traits and motives of cybercriminals. In computer science, cybersecurity aims to safeguard technology within computer systems, implementing security measures to prevent risks and threats that could harm the system. This field regulates security measures to thwart third-party invaders or intruders who engage in malicious activities such as stealing private, business, or organizational information for personal gain \cite{weimann2004cyberterrorism,li2021comprehensive,cremer2022cyber}.

In the domain of cybercrime, understanding the identity and motives of intruders plays a key role in mitigating risks to information security \cite{mcbrayer2014exploiting,kumar2016approaches,li2017review,ablon2018data,bada2020social,hunter2021factors,chng2022hacker,thackray2016social}. Psychological profiling emerges as a valuable tool for understanding the psychological traits and characteristics of cybercriminals, which strengthens strategies against potential cyber threats and assists in the identification of intruders and their motives through an examination of behavior, nature, and thought process.

Profiling in cybersecurity involves diverse criminological and criminal-law-based components, encompassing personal traits, criminal expertise, social attributes, and motivational factors. These elements help in understanding the predispositions, personality traits, demographics, socio-economic status, and motivations of cybercriminals, including those who are particularly elusive \cite{hani2024psychological,holt2024assessing}.

Cybercriminals frequently exhibit a range of psychological traits that strongly shape their behaviors and actions \cite{bada2020social,chng2022hacker,montanez2020human}. These individuals often possess a strong command of cyber technology, which they exploit for harmful purposes and various motives; common motives include financial gain, as seen in activities such as data theft and other forms of cyber fraud \cite{li2017review,holt2021examining}. Many are driven by greed, pursuing financial rewards, while others seek power or revenge against certain groups or institutions. Some cybercriminals are thrill-seekers, relishing the risk involved in their illicit activities, or opportunists who take advantage of vulnerabilities for personal benefit \cite{thackray2016social,saroha2014profiling,mcbrayer2014exploiting}. There are also those who simply disregard legal and ethical standards, compromising their reputations within the cyber community. Traits of fearlessness, with little regard for potential consequences, and a lack of empathy are also prevalent. Moreover, some individuals demonstrate boldness, testing their hacking abilities against individuals and organizations. Collectively, these traits paint a complex picture of the motivations and behaviors driving cybercriminals in various scenarios \cite{thackray2016social,li2017review,madarie2017hackers,chng2022hacker,maalem2020review}.

Motivating factors behind cybercriminal personality traits include revenge and blackmailing. Understanding these traits can help minimize security risks and enable better analysis and resolution of cybercrimes \cite{kipane2019meaning}. In addition, integrating findings from Large Language Models (LLMs) and psycholinguistic tools, such as the Linguistic Inquiry and Word Count (LIWC) dictionary and the Medical Research Council (MRC) psycholinguistic database \cite{ke2024exploring,boyd2022development,coltheart1981mrc}, into psychological profiling can significantly enrich the understanding of cybercriminal behaviors and motivations. This holistic approach to psychological profiling can not only reveal the complex personalities of cybercriminals but also strengthen overall security measures, protecting both individuals and organizations from cyber threats. In this paper, we explore the intersection of psychology and cybersecurity, with a specific emphasis on the role of LLMs and psycholinguistic features in profiling cyber threats.}

The remainder of this work is organized as follows. Section \S\ref{psyprofiling} discusses the fundamental role of psychological profiling in cybersecurity, outlining how it aids in understanding and mitigating the behaviors of cybercriminals. Section \S\ref{llmspsypro} explores the application of LLMs in psychological profiling, highlighting their potential to decode complex patterns of cybercriminal activity. In Section \S\ref{psyfeatures}, we examine the incorporation of psycholinguistic features into cybersecurity strategies, demonstrating how these tools can enhance the precision of psychological profiles. Section \S\ref{pfdiscussion} discusses different perspectives on psychological profiling in cybersecurity. Section \S\ref{ehtconspriv} addresses the ethical considerations and privacy implications inherent in the use of psychological profiling and data analysis in cybersecurity. Finally, Section \S\ref{fdfuturedirection} discusses future directions for research in this area and Section \S\ref{fdconclusion} concludes the paper with reflections on the evolving landscape of cybersecurity profiling.

\section{Psychological Profiling in Cybersecurity}\label{psyprofiling}

Researchers and practitioners reveal a complex profile of cyber criminals, showcasing  traits such as tech-savvy, well-networked, vengeful, goal-oriented, greedy, manipulative, risk-takers, opportunists, rule-breakers, fearless, emotionless, and daring \cite{mcbrayer2014exploiting,palassis2021exploration,saroha2014profiling,thackray2016social,li2017review,yang2018potential,holt2021examining}. More specifically, \citet{saroha2014profiling} identified a range of characteristics including smartness, creativity, and a need for control, shedding light on the multifaceted nature of individuals involved in cyber crimes, and uncovering motivating factors like monetary gain, thrill-seeking, and political beliefs that drive individuals towards engaging in cyber criminal activities.

In addition to profiling traits, understanding the psychological effects of cybercrime remains essential. \citet{gross2016psychological} indicated that exposure to cyber terrorism triggers heightened levels of stress and anxiety among individuals, akin to the psychological effects of conventional terrorism, emphasizing the pivotal role of perceived threats in shaping individuals' attitudes towards government surveillance, regulation, and military responses in the face of cyber threats. \citet{curtis2023understanding} underscored the significant influence of law enforcement's lack of cybercrime knowledge on low conviction rates and victim underreporting. The study revealed that victims often delay reporting cybercrimes due to embarrassment or a perception that they are better equipped to handle the situation themselves. This highlights the importance of training officers to increase their preparedness in dealing with cybercrime cases and engaging with victims.

In a related vein, \citet{palassis2021exploration} explored the psychological impacts of hacking victimization and underlined the need for support organizations to address these issues. The study underscores the importance of raising awareness about the psychological effects of cybercrime and promoting support opportunities for victims. Its findings provide valuable insights for clinicians and support organizations, informing the development of treatment guidelines and interventions to address the negative psychological impacts of hacking. \citet{gomez2018fear} investigated how limited experience and domain knowledge in cyberspace lead to the use of cognitive shortcuts and inappropriate heuristics, resulting in elevated levels of dread.

In recent investigations, building upon prior research, \citet{geer2023using} highlighted the importance of leveraging cybercriminals' cognitive biases to influence their behaviors during attacks. The study suggested that by using algorithms informed by cyberpsychology research, defenders can present low-risk, low-reward targets to steer hackers away from high-value assets. Studies show that attackers exhibit risk-averse behavior, preferring attacks on less secure machines to avoid the appearance of failure. Research on human subjects engaging in cybercriminal behavior revealed a strong relationship between key risk-taking and cybercriminal behaviors. \citet{bolton2019media} indicated that participants' exposure to fictional media, particularly crime-related television shows, can influence their attitudes towards criminal investigations and profiling techniques. The study revealed a correlation between media consumption habits and the perceived realism of investigative procedures portrayed in television episodes. Additionally, participants' beliefs about the role of criminal profilers and the importance of intuition in investigations were influenced by their media exposure. This underscores the nuanced relationship between media consumption and perceptions of criminal behavior and profiling accuracy.

Expanding upon the evolving understanding of cybercriminal behavior, \citet{lickiewicz2011cyber} highlighted the significance of intelligence, personality traits, and social skills in the effectiveness of cyber attacks. The study emphasized the role of environmental factors, such as family relationships and educational background, in shaping the behaviors of hackers. It suggested that a holistic approach, considering both individual characteristics and external influences, is crucial for developing a comprehensive psychological profile of cyber criminals. Additionally, the study noted the need for interdisciplinary collaboration between information technology and investigative psychology to combat cybercrime.

Psychological profiling, rooted in behavioral analysis and psychological theory, aims to uncover patterns and traits indicative of malicious intent in cyber activities. This approach utilizes various aspects of human behavior, such as language use, decision-making processes, and emotional responses, to discern the psychological profiles of threat actors \cite{thackray2016social,jiang2018prediction,kipane2019meaning,hani2024psychological,budimir2021emotional,bada2021profiling,montanez2020human,gaia2020psychological,zambrano2023modeling,kioskli2022estimating}. Leveraging techniques from psychology, including personality assessment and psycholinguistic analysis, enables the identification of anomalous behaviors and potential indicators of cyber threats. 

For instance, \citet{kioskli2022estimating} emphasized the importance of profiling potential attackers in cybersecurity to enhance the accuracy of vulnerability severity scores using psychological and behavioral traits. Research investigated the influence of cultural and psychological factors on cyber-security behavior, utilizing the Big Five Framework to assess personality traits and their impact on user attitudes towards privacy and self-efficacy \cite{halevi2016cultural,odemis2022detecting}. {\color{black}More specifically,} \citet{hani2024psychological} proposed machine learning models for psychological profiling of hackers based on the ``Big Five'' personality traits model (OCEAN - Openness, Conscientiousness, Extroversion, Agreeableness, Neuroticism) and their models achieved 88\% accuracy in mapping personality clusters with different types of hackers (White Hat, Grey Hat, etc.), identifying cyber-criminal behaviors. \citet{gaia2020psychological} discovered that individuals attracted to hacking exhibit high scores on Machiavellianism and Psychopathy scales, with Grey Hat hackers showing opposition to authority, Black Hat hackers scoring high on thrill-seeking, and White Hat hackers displaying tendencies towards Narcissism. The Dark Triad traits significantly predict interest in different types of hacking, while thrill-seeking emerges as a key motivator for Black Hat hackers. Perceptions of apprehension for violating privacy laws negatively impact Grey Hat and Black Hat hacking.

\begin{table*}[h!]
\fontsize{8.25}{8.25}\selectfont
\centering
\caption{Summary of LLM applications in psychological profiling in cybersecurity}\label{summary_llms}
\renewcommand{\arraystretch}{1.6} % Adjust the row height
\begin{tabular}{|C{2.5cm}|C{4cm}|C{4.2cm}|C{3.5cm}|}
\hline
\textbf{Research} & \textbf{Focus} & \textbf{Cybersecurity applications} & \textbf{Sources of data} \\
\hline
\citet{petrov2024limited} & Simulating human psychological behaviors using LLMs & Evaluating psychometric properties for profiling potential threats & Standardized personality constructs \\
\hline
\citet{pellert2023ai} & Repurposing psychometric inventories for LLMs & Profiling values, morality, and beliefs to detect radicalization & Standard psychometric inventories \\
\hline
\citet{sorokovikova2024llms} & Fine-tuning LLMs on Big Five traits & Profiling based on language to identify potential threats & Psychometric test items \\
\hline
\citet{safdari2023personality} & Administering personality tests on LLMs & Mimicking specific human personality profiles for threat detection & Personality tests \\
\hline
\citet{huang2023humanity} & PsychoBench framework for evaluating LLM personalities & Understanding complex psychological profiles for enhanced cybersecurity & Personality traits, interpersonal relationships, motivational tests, emotional abilities \\
\hline
\citet{frisch2024llm} & Conditioning LLM agents on personality profiles & Mimicking human traits for improved phishing and social engineering detection & Persona conditioning data \\
\hline
\citet{yamin2021weaponized} & Weaponized use of LLMs in cyber attacks & Generating malicious code, automated hacking, phishing & Training data on malware and exploits \\
\hline
\citet{motlagh2024large} & Generating malicious payloads with LLMs & Creating new strains of malware & Relevant cybersecurity data \\
\hline
\citet{beckerich2023ratgpt} & Using LLMs for automated hacking & Vulnerability scanning and developing exploits & Hacking toolkits \\
\hline
\citet{schmitt2023digital} & Social engineering and phishing & Mimicking human language for cyber attacks & Historical phishing data \\ \hline

\citet{zhang2024psysafe} & PsySafe for framework understanding and mitigating risks arising from dark psychological states & Identifying vulnerabilities, evaluating safety, and implementing defense mechanisms  & Psychological assessments, behavioral evaluations\\  %\hline

%\citet{kioskli2022estimating} & \textcolor{purple}{ Optimisation of vulnerability severity scores by integrating attacker profiles} & Customisation of security measures, Improved organisational resilience, Inclusion of human factors in risk assessment & Psychological assessments, behavioral evaluations\\

\hline
\end{tabular}
\end{table*}

Moreover, \citet{kipane2019meaning} revealed that cybercriminals exhibit a range of behaviors and traits that deviate from societal norms, influenced by factors such as heredity, education, culture, and socio-economic status. Profiling methods focus on identifying key psychological features, modus operandi, and criminal motivations to aid in early detection and investigation of cybercrimes. The study emphasizes the significance of expert knowledge and advanced technologies in enhancing law enforcement efforts to combat cybercrime. Overall, the research underscores the evolving nature of criminal profiling in the digital era and the critical role it plays in addressing the growing threat of cybercriminal activities. In response to the escalating threat posed by cybercrimes, \citet{thackray2016social} highlighted the diverse motivations of hackers, including recreation, prestige, revenge, profit, and ideology, which influence their engagement in cyber activities. The study underscores the importance of not only teaching coding skills but also educating individuals about the risks and consequences of online actions to prevent cyber-crime involvement. Additionally, the research emphasizes the need to identify at-risk groups and individuals to target awareness campaigns and promote informed online behavior for future generations. Lastly, the study suggests that understanding social psychological theories can enhance communication with hacker communities and individuals, ultimately contributing to more effective cybersecurity practices.

\section{LLMs in Psychological Profiling}\label{llmspsypro}
Large Language Models (LLMs), such as OpenAI's GPT series of models, Google's PaLM and Gemini, and Meta's LLaMA family of open-source models, have demonstrated remarkable capabilities in natural language understanding and generation tasks \cite{minaee2024large}. As these models continue to evolve and become more sophisticated, researchers and practitioners are exploring their potential applications beyond language tasks, venturing into the realm of psychological profiling (see Table \ref{summary_llms}). These models are utilized to profile individuals based on their language use patterns and communication styles, facilitating the early detection of potential threats \cite{ke2024exploring}.

The potential applications of LLM-based psychological profiling are vast and diverse \cite{abdurahman2024evaluating,ke2024exploring,hani2024psychological,pellert2023ai,petrov2024limited,huang2023humanity}. In mental health settings, these techniques aid in the early detection of psychological disorders and the development of personalized treatment plans \cite{lai2023psy,chung2023challenges,hagendorff2023machine}. In human-AI interaction, understanding the perceived personalities of LLMs improves user engagement and trust, leading to more natural and effective interactions \cite{sharma2024investigating}.

However, the application of LLMs to psychological profiling is not without challenges and ethical considerations. Existing personality models and assessment methods have been developed primarily for human subjects, and their suitability for evaluating artificial intelligence systems is questionable. Additionally, the fluid and context-dependent nature of LLM ``personalities'' raises concerns about the reliability and validity of traditional personality assessment techniques when applied to these models \cite{sorokovikova2024llms}. As researchers delve deeper into this emerging field, they must grapple with the complexities of transferring human-centric concepts like personality to artificial intelligence systems. LLMs are explored for psychological profiling tasks, such as detecting personality traits, values, and other non-cognitive characteristics \cite{hani2024psychological,frisch2024llm,pellert2023ai,petrov2024limited,huang2023humanity,song2024identifying,safdari2023personality,hani2024psychological,sorokovikova2024llms,zhang2024psysafe}. 

In exploring the multifaceted landscape of psychological profiling with LLMs, researchers have embarked on various avenues to understand their potential applications. For instance, \citet{petrov2024limited} focused on investigating the ability of LLMs to simulate human psychological behaviors using prompts to adopt different personas and respond to standardized measures of personality constructs to assess their psychometric properties. \citet{pellert2023ai} repurposed standard psychometric inventories originally designed for assessing human psychological characteristics, such as personality traits, values, morality, and beliefs, to evaluate analogous traits in LLMs. \citet{sorokovikova2024llms} fine-tuned LLMs on psychometric test items related to the Big Five personality traits for evaluating personalities based on language. \citet{safdari2023personality} introduced a method for administering personality tests on LLMs and shaping their generated text to mimic specific human personality profiles.

Furthermore, \citet{huang2023humanity} proposed PsychoBench, a framework for evaluating personality traits, interpersonal relationships, motivational tests, and emotional abilities to uncover complex psychological profiles within LLMs and their potential integration into human society as empathetic and personalized AI-driven solutions. \citet{frisch2024llm} demonstrated that LLM agents conditioned on personality profiles can mimic human traits, with creative personas displaying more consistent behavior in both interactive and non-interactive conditions; the research highlights the importance of robust persona conditioning in shaping LLM behavior and emphasizes the asymmetry in linguistic alignment between different persona groups during interactions. 

\citet{zhang2024psysafe} presented PsySafe, a framework designed to evaluate and improve the safety of multi-agent systems (MAS) by addressing the psychological aspects of agent behavior. PsySafe incorporates dark personality traits to assess and mitigate potential risks associated with agent behaviors in MAS; in addition, it includes identifying vulnerabilities, evaluating safety from psychological and behavioral perspectives, and implementing effective defense strategies. The findings yielded by PsySafe reveal several phenomena, including collective dangerous behaviors among agents, their self-reflection on engaging in such behaviors, and the correlation between psychological assessments and behavioral safety.

While LLMs offer promising applications in psychological profiling, their language generation capabilities also raise concerns about potential misuse for cyber attacks and malicious activities \cite{yamin2021weaponized,motlagh2024large,gupta2023chatgpt,yao2024survey}. Attack payloads and malware creation involve LLMs generating malicious code or new strains of malware through training on relevant data \cite{beckerich2023ratgpt,wu2023deceptprompt}. Automated hacking and vulnerability scanning tasks can be performed by LLMs, including generating code for automated hacking attacks, scanning software for vulnerabilities, or developing exploits \cite{wu2023deceptprompt,xu2024autoattacker}. 

In addition, LLMs can be used for social engineering and phishing purposes, leveraging their ability to mimic human language patterns to create convincing social engineering attacks, phishing emails, or disinformation campaigns \cite{schmitt2023digital}. Adversaries could potentially manipulate LLM outputs for malicious purposes using prompt injection techniques \cite{liu2023prompt,piet2023jatmo}. LLMs can generate highly personalized and persuasive phishing emails tailored to specific individuals within an organization, bypassing traditional detection systems. Studies show these AI-crafted attacks can be strikingly effective, with around 10\% of recipients entering credentials on fake login portals \cite{bethany2024large}. The ability of LLMs to mimic human language patterns and adapt to different contexts makes them a powerful tool for deception and manipulation \cite{prome2024deception}.

The 2023 Report of Voice of SecOps provides a comprehensive analysis of threats and stressors posed by LLMs, revealing that 51\% of security professionals are likely to leave their job within 2024.\footnote{\textit{Generative AI and Cybersecurity: Bright Future or Business Battleground?} Deep Instinct. (2023). Voice of SecOps Reports. Retrieved from \url{https://www.deepinstinct.com/voice-of-secops-reports}. Accessed on May 12, 2024.} The study surveyed over 650 senior security operations professionals in the U.S. to assess LLMs' impact on the cybersecurity industry. Findings indicate a 75\% surge in attacks in 2022, with 85\% attributing this increase to bad actors leveraging LLMs. Furthermore, 70\% of respondents believe LLMs positively influence employee productivity and collaboration, while 63\% perceive an enhancement in employee morale. Ransomware emerges as the greatest threat to organizational data security, with 46\% of respondents acknowledging its severity and 62\% indicating it as the top C-suite concern, a notable increase from 44\% in 2022; the pressure to combat ransomware has prompted organizations to revise their data security strategies, with 47\% now possessing a policy to pay the ransom, compared to 34\% in the previous year. Moreover, the report reveals a 55\% increase in stress levels among security professionals, primarily attributed to staffing and resource constraints, cited by 42\% of respondents.

\section{Psycholinguistic Features}\label{psyfeatures} 

Psycholinguistic features encompass a wide range of linguistic attributes and psychological constructs that reflect cognitive and emotional aspects of language use. Integrating psycholinguistic features into cybersecurity frameworks enhances the granularity of threat profiling techniques and enables a deeper understanding of cybercriminals' mental states and feelings \cite{jiang2018prediction,deb2018predicting,uyheng2022language,krylova2019psycholinguistic,xu2023determining}. Psycholinguistic features include sentiment analysis, linguistic complexity measures, lexical diversity metrics, and stylistic characteristics. Through advanced text analysis algorithms and machine learning algorithms, these features can be leveraged to identify anomalous patterns indicative of malicious intent.

One of the powerful tools in psycholinguistic analysis is the Linguistic Inquiry and Word Count (LIWC) dictionary \cite{boyd2022development}. In the context of cyber attacks, LIWC has been used to detect deception in phishing emails by analyzing the psycholinguistic features that attackers employ to deceive end-users \cite{xu2023determining}. Research shows that phishers often use language conveying certainty (e.g. always, never), time pressure and work-related words to increase vulnerability of targets. Conversely, reward-related words like money or cash tend to decrease vulnerability as they are associated with scams. Beyond phishing, LIWC has been applied to study online predator behavior, analyze developer personalities, model social media rumors, and understand user reactions in crowdsourcing \cite{rogers2006self,tausczik2010psychological,shappie2020personality,kranenbarg2023there,budimir2021emotional}.

Building on the potential of LIWC for psycholinguistic analysis in cybersecurity, researchers explore its applications to understand attacker behavior and victim vulnerabilities. More precisely, \citet{guo2023text} focused on analyzing the vulnerability factors of potential victims to cybergrooming using LIWC to quantify and understand the social-psychological traits that may make individuals more susceptible to online grooming; they reveal significant correlations between specific vulnerability dimensions and the likelihood of being targeted as a victim of cybergrooming. Interestingly, the research observed negative correlations between victims and certain family and community-related traits, challenging conventional beliefs about the key factors contributing to vulnerability in online contexts. \citet{na2019unified} utilize LIWC and demonstrate that malicious insiders exhibit specific linguistic patterns in their written communications, including increased use of self-focused words, negative language, and cognitive process-related words compared to other team members; as insiders become more detached from the team, language similarity decreases over time.

In a different angle, psycholinguistic features were utilized to examine the manipulative aspects of cybercrimes. More specifically, \citet{krylova2019psycholinguistic} investigated the psycholinguistic dimensions of social engineering within cybersecurity, employing activity theory to dissect the methods and techniques utilized by malicious actors. This research reveals the sophisticated tactics employed by social engineers to manipulate emotions, impede critical thinking, and exploit moral values to influence user behavior and extract sensitive information. \citet{parapar2014combining} proposed a machine learning model for detecting sexual predation in chatrooms using psycholinguistic, content-based, and chat-based features, and show distinct characteristics that differentiate predators from non-predators. Particularly, \citet{rogers2006self} investigated the psychological traits and behaviors of individuals involved in self-reported criminal computer activities, emphasizing the role of extraversion in predicting such behavior and challenging stereotypes by shedding light on the complexities of personality factors in criminal/deviant computer behavior through the use of Likert-scale questionnaires and psychometric instruments.

Furthermore, \citet{chatterjee2021vulnerable} conducted a study on phishing influence detection using a novel computational psycholinguistic analysis approach to identify influential sentences that could potentially lead to security breaches and hacking in online transactions and social media interactions, developing a language and domain-independent computational model based on Cialdini's principles of persuasion.\footnote{\textit{The 6 Principles of Persuasion: Tips from the leading expert on social influence}, Douglas T. Kenrick. Posted Dec. 8, 2012. Retrieved from \url{https://www.psychologytoday.com/ca/blog/sex-murder-and-the-meaning-of-life/201212/the-6-principles-of-persuasion}. Accessed May 20, 2024.} \citet{kranenbarg2023there} indicated that cyber offenders displayed similarities to the community sample on certain traits but exhibited differences from offline offenders, particularly in conscientiousness and openness to experience. Notably, cyber offenders showed lower scores on honesty-humility compared to the community sample, suggesting potential implications for intervention strategies targeting specific personality traits in this population.

\citet{budimir2021emotional} emphasized the importance of understanding psycholinguistic features and psychology in cybersecurity to develop effective strategies and interventions. They explore the emotional responses triggered by cybersecurity breaches, focusing on the hacking of smart security cameras. The study identifies a 3-dimensional structure of emotional reactions, highlighting negative affectivity, proactive versus fight/flight action tendencies, and emotional intensity and valence. Personality characteristics, such as the Big Five traits and resilient/overcontrolled/undercontrolled types, were found to relate to these emotional dimensions.

Recently, the application of sentiment analysis techniques has paved the way for building psychological profiles and detecting and understanding cyber threats. \citet{sapienza2017early} utilized sentiment analysis to identify discussions around exploits, vulnerabilities, and attack planning on dark web forums even before these threats manifest in the real world, and to provide early warnings through the observation of changes in sentiment and semantic context. \citet{deb2018predicting} proposed approaches to predict cyber-events by leveraging sentiment analysis on hacker forums and social media to analyze the sentiment expressed in online discussions and detect signals that may precede cyber attacks. \citet{jiang2018prediction} built user psychological profiles based on the sentiment analysis of their network browsing and email content, and demonstrate that this approach can proactively and accurately detect malicious insiders with extreme or negative emotional tendencies.

Building upon recent studies and advancements, \citet{uyheng2022language} developed a machine learning model called TrollHunter and collected a dataset of online trolling messages and found that troll messages exhibit more abusive language, lower cognitive complexity, and greater targeting of named entities and identities; the model achieved an 89\% accuracy rate and F1 score in identifying trolling behavior.

\section{Discussion}\label{pfdiscussion}

The integration of psychological profiling into cybersecurity practices offers a multifaceted approach to understanding and mitigating cyber threats. LLMs and psycholinguistic features provide deeper understanding into the behaviors, motivations, and emotional states of cybercriminals. This discussion section explores the potential benefits, and challenges of these techniques, drawing from the research findings presented earlier.

\subsection{Benefits of Psychological Profiling in Cybersecurity}

Psychological profiling in cybersecurity holds significant promise. Identifying psychological traits and patterns in cybercriminal behavior enables security professionals to anticipate and preemptively counteract potential threats. For instance, understanding the personality traits and motivations of different types of hackers (e.g., White Hat, Black Hat, Grey Hat) allows for more tailored security measures and interventions  \cite{hani2024psychological,gaia2020psychological}. The use of LLMs enhances this profiling by analyzing large volumes of text data, identifying linguistic patterns that may indicate malicious intent.

Psycholinguistic features, such as those derived from the LIWC dictionary, provide additional granularity. These features help in detecting subtle cues in language that might indicate deception, stress, or malicious intent. For example, certain linguistic markers can distinguish phishing emails from legitimate communications, thereby improving the accuracy of threat detection systems \cite{xu2023determining,rogers2006self}.

Moreover, the incorporation of psychological profiling can aid in the development of more personalized cybersecurity training programs. Understanding the psychological traits that make individuals more susceptible to cyber attacks allows organizations to design targeted awareness campaigns and training modules that address specific vulnerabilities.

\subsection{Challenges and Limitations}

Despite the promising applications, several challenges and limitations need to be addressed. One major challenge is the accuracy and reliability of psychological profiling techniques. %While LLMs and psycholinguistic tools provide valuable information, they are not infallible. False positives and negatives can occur, leading to either unnecessary alarm or undetected threats. Ensuring the robustness and validity of these models is crucial for their effective deployment in real-world scenarios \cite{xu2023determining,hani2024psychological}.

While LLMs and psycholinguistic tools provide valuable insights, they come with inherent limitations. Implementing and maintaining these advanced profiling systems require a workforce equipped with specialized skills in artificial intelligence, cybersecurity, and psychological analysis. There is often a shortage of professionals with the necessary expertise to develop, deploy, and refine these tools. Addressing this skill gap is crucial for the effective utilization of psychological profiling in cybersecurity.

The effectiveness of LLMs largely depends on the quality and diversity of the data they are trained on. Inaccurate models can result from poor-quality data, such as poisoned or contaminated datasets, or from non-representative data. Moreover, acquiring diverse and representative datasets is particularly challenging in the field of cybersecurity, where data sensitivity and proprietary information are significant concerns.

Additionally, the use of these tools can lead to false positives and negatives, causing either unnecessary alarms or undetected threats. Thus, ensuring the robustness and validity of these models is vital for their successful deployment in real-world scenarios \cite{xu2023determining,hani2024psychological}.
 
Another challenge lies in the dynamic and evolving nature of cybercriminal behavior. Cybercriminals continually adapt their tactics to evade detection, which means that profiling techniques must also evolve. Continuous updates and refinements to the models and algorithms are necessary to keep pace with these changes.

The ethical implications of psychological profiling in cybersecurity cannot be overlooked. The use of personal data to create psychological profiles raises significant privacy concerns. It is essential to balance the benefits of enhanced security with the protection of individual privacy rights. Transparent policies and stringent data protection measures must be in place to ensure that the use of psychological profiling does not infringe on personal freedoms.

\section{Ethical Considerations}\label{ehtconspriv}
Ethical considerations are paramount when employing psychological profiling in cybersecurity. The potential for misuse of these technologies for surveillance, manipulation, or discrimination is a serious concern. For example, the ability of LLMs to generate persuasive phishing emails tailored to specific individuals poses a significant threat if used maliciously \cite{liyanage2023ethical}.

To mitigate these risks, it is crucial to establish ethical guidelines and regulatory frameworks that govern the use of psychological profiling tools. These guidelines should emphasize the importance of informed consent, data minimization, and transparency in the use of personal data. Additionally, there should be mechanisms for accountability and oversight to ensure that these technologies are used responsibly and ethically \cite{mcstay2020emotional,fleming2021considerations}.

\section{Future Directions} \label{fdfuturedirection}

Future research should focus on improving the robustness of psychological profiling techniques. This includes developing more sophisticated models that can adapt to the evolving tactics of cybercriminals and integrating multimodal data sources (e.g., text, behavioral data, biometric data) to create more comprehensive profiles.

Another promising direction is the exploration of collaborative approaches that combine human expertise with machine intelligence. Human analysts and AI systems can collaborate to achieve more effective and nuanced threat detection and mitigation strategies.

Finally, ongoing efforts to address the ethical and privacy concerns associated with psychological profiling are essential. This includes developing new methods for anonymizing and protecting personal data while still enabling meaningful analysis, as well as fostering a culture of ethical awareness and responsibility among cybersecurity professionals.

\section{Conclusion}\label{fdconclusion}
The integration of psychological profiling, LLMs, and psycholinguistic features into cybersecurity practices represents a significant advancement in the field. These techniques offer the potential to enhance threat detection and mitigation strategies by providing deeper understanding into the behaviors and motivations of cybercriminals. However, realizing this potential requires addressing the challenges and ethical considerations associated with these technologies. By doing so, we can create more robust and responsible cybersecurity frameworks that protect both organizations and individuals from evolving cyber threats.

\section*{Acknowledgments}
The authors thank all Greprovad members for helpful discussions and comments on early drafts.

\bibliography{custom}
\bibliographystyle{acl_natbib}

\end{document}